\documentclass{article}

\PassOptionsToPackage{numbers, compress}{natbib}

\usepackage[numbers]{natbib}
\usepackage[final]{neurips_2024}
\usepackage{minted}
\usepackage{subcaption}
\usepackage{graphicx} 
\usepackage{array} 
\usepackage{booktabs}
\usepackage[utf8]{inputenc}
\usepackage{tabularx}

\usepackage{amsmath}

\usepackage[utf8]{inputenc} 
\usepackage[T1]{fontenc}    
\usepackage{hyperref}       
\usepackage{url}            
\usepackage{booktabs}       
\usepackage{amsfonts}       
\usepackage{nicefrac}       
\usepackage{microtype}      
\usepackage{xcolor}         

\title{Final Report - Fine Tuning Methods for Low-resource Languages}

\author{%
  \begin{minipage}{0.45\textwidth}
    \centering
    \textbf{Anton Johansson} \\
    \normalfont 
    Department of Computer Science,\\
    Tsinghua University\\
    \texttt{wh24@mails.tsinghua.edu.cn}
  \end{minipage}
  \hfill 
  \begin{minipage}{0.45\textwidth}
    \centering
    \textbf{Tim Bakkenes} \\
    \normalfont 
    Department of Computer Science,\\
    Tsinghua University\\
    \texttt{dim24@mails.tsinghua.edu.cn}
  \end{minipage}
  \vspace{0.5cm} 
  \\
  \begin{minipage}{\textwidth}
    \centering
    \textbf{Daniel Wang} \\
    Department of Electronic Engineering,\\
    Tsinghua University\\
    \texttt{wyl24@mails.tsinghua.edu.cn}
  \end{minipage}
}

\begin{document}

\maketitle

\vspace{-0.5cm}
\begin{abstract}
The rise of Large Language Models has not been inclusive of all cultures. The models are mostly trained on English texts and culture which makes them underperform in other languages and cultural contexts. By developing a generalizable method for preparing culturally relevant datasets and post-training the Gemma 2 model, this project aimed to increase the performance of Gemma 2 for an underrepresented language and showcase how others can do the same to unlock the power of Generative AI in their country and preserve their cultural heritage.

\end{abstract}

\section{Introduction}
In this report, the final project of the Fall 2024 Machine Learning course is described and discussed. This project was inspired by the Kaggle competition \textit{Google - Unlock Global Communication with Gemma}, which aims to improve the performance of the Gemma 2 LLM for underrepresented languages as an innovative solution for global communication \cite{kaggle_gemma_comp}. This competition is in line with the UNESCO initiative \textit{'Language Technologies for All'}, ensuring that AI models support a broad range of languages, which is vital to preserve global cultural heritage and prevent digital language extinction \cite{lt4all2025}.

\subsection{Background}

The fast progress in Artificial Intelligence (AI) means that more and more people are using Large Language Models (LLMs). Since most LLMs are trained primarily on English data, other languages and cultures risk fading away, a concern highlighted by the World Economic Forum\cite{wef_2024}. As these biased models become increasingly integrated into society, their potential to amplify existing biases and generate harmful content is intensified, as noted by \citet{arxiv_2410_13868}. Also, it contributes to the “digital language death,” where minority languages risk losing their online presence according to Soria et al. \cite{soria2020linguistic}. This emphasizes the importance of developing LLMs that support more than just the English language and culture. Despite some multilingual capabilities, models like Gemma often underperform in non-English languages, showing reduced accuracy and consistency \cite{wef_2024}.




\subsection{Importance and Impact}

It has been established that Google's Gemma 2 model was primarily trained on English data, leading to reduced performance in other languages and cultures. According to \citet{arxiv_2312_01509}, the ideal scenario would be to design LLMs with underrepresented groups in mind from the start. However, even though this might be possible in the future, the bias problem needs to be addressed immediately.

By showcasing methods and technologies whilst developing a fine-tuned version of Gemma 2, other developers will have a roadmap to adapt Gemma 2 for their purposes. Tailoring LLMs to underrepresented languages will empower people from  all around the globe to utilize LLMs to solve diverse real-world problems. At the same time, it can be used to preserve the linguistic and cultural identities of local communities and enhance global communication. 

Further highlighting the importance of their project, recent efforts such as BigScience’s BLOOM \cite{bigscience2022bloom} and Meta’s “No Language Left Behind” \cite{meta2022nllb} illustrate the growing focus on multilingual AI. These open-source projects highlight the urgency of addressing language coverage gaps. Also, Europe’s eTranslation platform \cite{europe2022etranslation} and UNESCO’s digital preservation efforts \cite{lt4all2025} emphasize the global demand for robust multilingual technologies. 

However, the applications of tailoring LMMs is not limited to language and culture. The methods proposed could also be used to tailor models for specific business related tasks \cite{google_gemma_docs}. Furthermore, the model developed during this project could enable non-English speakers to utilize a LLM.



\section{Formal Problem Definition}
The goal is to fine tune googles' pre-trained Gemma 2 model for other languages than English and showcase and document the methods used to achieve this. In order to do this, a processed dataset used for post-training to fine tune the model, $D^{\prime}$, has to be crafted from various sources. The Gemma 2 model parameters, $\theta$, and the context supplied via features like Retrieval-Augmented-Generation, $C_{\text{LLM}}$, should be optimized to minimize the loss function that can be expressed with the following equation where $y$ is the expected result and $M_{\text{Gemma2}}(x, C_{\text{LLM}}; \theta)$ is the predicted result of the model given the prompt $x$, the context $C_{\text{LLM}}$ and the model parameters $\theta$.
\[
\sum_{(x, y) \in D'} L(y, M_{\text{Gemma2}}(x, C_{\text{LLM}}; \theta))
\]
In other words, the goal is to find the optimal model parameters, $\theta'$, and the best context generation to provide the optimal context, $C_{\text{LLM}}'$, that minimizes the loss function. The problem is expressed in the equation below.
\[
(\theta', C_{\text{LLM}}') = \arg \min_{\theta, C_{\text{LLM}}} \sum_{(x, y) \in D'} L(y, M_{\text{Gemma2}}(x, C_{\text{LLM}}; \theta))
\]

The explanation above is a general description for the underlying optimization problem that we have worked on. However, in reality it is not clear what a good response is when it comes to language, historical facts and cultural relevance. The goal could therefore be augmented by explaining how a \textit{good response} can be defined in terms of cultural authenticity, factual accuracy, and historical relevance. These criteria could for example be integrated into the training and evaluation by adding human feedback. Because of this, the language that was chosen for this project was Swedish since two thirds of the group members have native proficiency in the language.

\section{Related Work For Inspiration}
Recently \citet{huggingface_gemma_jpn} has fine-tuned Gemma 2 for Japan. They used reinforcement learning from human feedback and specialized post-training techniques to enhance performance in Japanese, without impacting the English performance. 
Furthermore, \citet{gpt_sweden} has developed a Swedish large-scale generative language model to help Swedish organizations build language applications that previously were not possible. To do this, AI Sweden had to source Swedish data to train the model and documented it well which makes their work relevant to this project.

Besides Google’s work on Gemma 2 and AI Sweden’s Swedish model, projects previously mentioned such as BLOOM \cite{bigscience2022bloom} and Meta’s “No Language Left Behind” \cite{meta2022nllb} have released extensive open-source multilingual datasets and training pipelines. These resources can inform best practices for those seeking to adapt Gemma 2 to underrepresented languages.



\section{Methodology}

This section will cover the methodology for completing this project. There were many parts to consider. For example, a dataset had to be crafted to fine tune the model for it to learn how to write better in Swedish. Another dataset for giving the model external data about Swedish culture and history also had to be curated. Then the model also had to be fine tuned and the mechanics for retrieving relevant context given a prompt also had to be understood. 

\begin{figure} [H]
    \centering
    \includegraphics[width=0.5\linewidth]{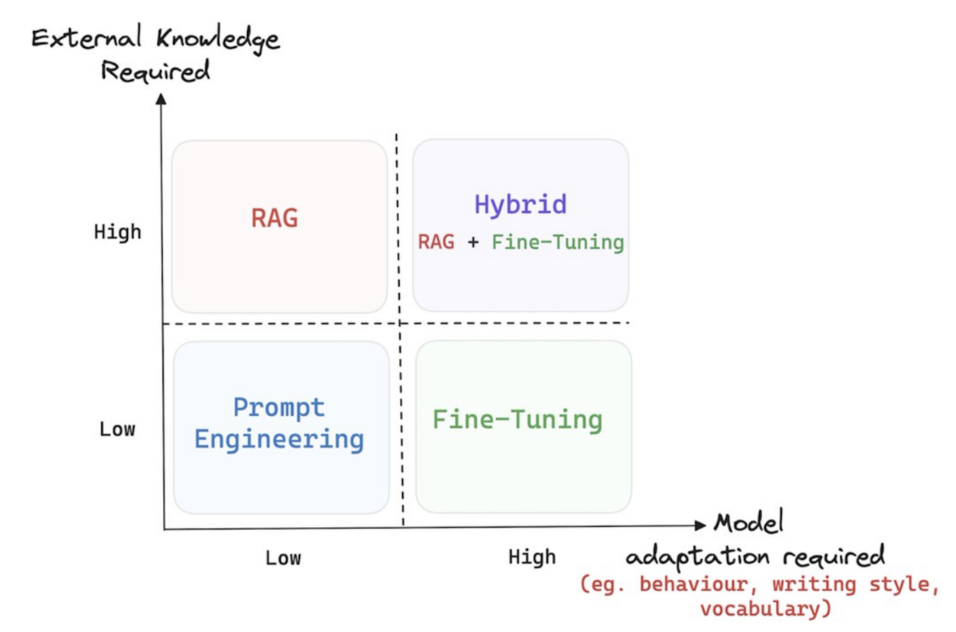}
    \caption{Method Matrix \cite{MLSpring2024}}
    \label{fig:method-matrix}
\end{figure}

As can be seen in figure \ref{fig:method-matrix}, a hybrid approach was used in this project where both external knowledge and model adaptation to a specific language was required. These topics are covered in more detail in the following subsections.

\subsection{Datasets}
There were two main parts to this project, fine tuning Gemma 2 for Swedish language and enriching its responses by incorporating external data as context. This meant that two datasets had to be created. One fine tuning dataset and one retrieval augmented generation dataset.

\subsubsection{Fine Tuning Dataset}

Since Gemma 2B already has a basic understanding of Swedish, and given our computational resources are limited while the model itself is relatively small (2 billion parameters), we decided to fine-tune it using a dataset of 500 prompts and related answers. 

We began the fine-tuning process by curating a small dataset of 50 prompts, observing a slight performance boost. Encouraged by this, we expanded the dataset to 100 prompts, then to 200 prompts, and finally to 500 prompts. At this stage, we saw noticable improvement in performance. However even more prompts would likely improve performance even more but due to time and computational constraints we settled for 500 prompt/response pairs. The prompts were hand-crafted by us with assistance from ChatGPT. For generating these prompts, we used a pretrained Swedish language model available in the OpenAI store and prompted it to create prompts and answers in Swedish suitable for a large language model \cite{ChatGPT2024}. The prompts were designed to cover a diverse range of topics, including Swedish culture, the Swedish language, education in Sweden, translation tasks, and other commonly encountered subjects.

Once the prompts and answers were generated, each response was constrained to approximately 25 tokens in length. We then manually reviewed all the answers to ensure they were free from grammatical errors and factual inaccuracies. The model used to generate the prompt-answer pairs performed well, with an error rate of approximately one mistake per five answers.This systematic approach allowed us to fine-tune Gemma 2B effectively while ensuring high-quality data for the fine-tuning process.

In the table below, we present four examples of prompts and their corresponding answers in Swedish, along with their English translations for clarity and ease of understanding.

\begin{table}[H]
    \centering
    \caption{Prompts and Answers in Swedish and English}
    \small 
    \begin{tabularx}{\textwidth}{ 
        >{\raggedright\arraybackslash}X 
        >{\raggedright\arraybackslash}X 
        >{\raggedright\arraybackslash}X 
    }
        \toprule
        \textbf{Prompt (Svenska / English)} & \textbf{Svar (Svenska)} & \textbf{Answer (English)} \\
        \midrule
        Vad är typiskt för svensk midsommar? / What is typical of Swedish midsummer? 
        & Svensk midsommar är en av de mest älskade högtiderna. Den firas med dans runt midsommarstången, traditionella sånger och mat som sill, potatis, gräddfil och jordgubbar. Kvinnor bär kransar av blommor i håret och firandet sker ofta utomhus när vädret tillåter det, dock regnar det ofta på midsommar. Det är en högtid fylld med gemenskap och glädje.
        & Swedish midsummer is one of the most loved holidays. It is celebrated with dancing around the midsummer pole, traditional songs and food such as herring, potatoes, sour cream and strawberries. The women wear flowers in their hair and the celebration often takes place outdoors when the weather permits, however it often rains during midsummer. It is a holiday filled with community and joy.
        \\
        \midrule
        Hur fungerar Allemansrätten? / How does "Allemansrätten" work? 
        & Allemansrätten ger alla i Sverige rätt att vistas i naturen, oberoende om det är privat eller allmän mark. Man får plocka bär och svamp och bada i sjöar, så länge man inte stör eller förstör. Det är en unik del av svensk kultur och naturvård.
        & "Allemansrätten" gives everyone in Sweden the right to be in nature, regardless of whether it is private or public land. You can pick berries and mushrooms and swim in lakes, as long as you don't disturb or destroy the environment. It is a unique part of Swedish culture and nature conservation.
        \\
        \midrule
        Hur hanterar jag stress under tentaperioden? / How do I deal with stress during the exam period? 
        & Planera studierna i små block med pauser, motionera för att släppa stress och sov regelbundet. Öva avslappningsövningar som djupandning vid behov.
        & Plan your studies in small blocks with breaks, exercise to release stress and sleep regularly. Practice relaxation exercises such as deep breathing when needed.
        \\
        \midrule
        Vad är en synonym för ordet "fantastisk"? / What is a synonym for the word "awesome"? 
        & En passande synonym är "enastående" eller "underbar", beroende på sammanhang.
        & A suitable synonym is "outstanding" or "wonderful", depending on the context.
        \\
        \bottomrule
    \end{tabularx}
\end{table}

\subsubsection{Retrieval Augmented Generation Dataset}
In order to be able to do Retrieval Augmented Generation, henceforth referenced as RAG, a dataset must first be created that contains the knowledge that is wished to be known by the model. The task of creating a good dataset is difficult and often requires a lot of people and time. Because of this, the size of the dataset was kept small because of the limited time and number of people involved in this project. The RAG still managed to give meaningful results. There are many things to think about when creating a dataset from scratch. Below follows information about how it can be done and how it was done for this project.

\textbf{Identify sources and domain} 

First of all, feasible sources that can be used to collect data for the given domain has to be identified in order for the author of the dataset to find the relevant data. At the same time as identifying relevant sources, the domain that the system needs to have external data incorporated from also has to be identified. For this project the domain was Swedish culture and history and the main sources used was various museums in Sweden that have published information about the history and culture of Sweden throughout the years, Wikipedia\cite{SwedishWikipediaHuvudsida} articles about the culture of Sweden and an initiative called LitteraturBanken\cite{litteraturbanken}. LitteraturBanken\cite{litteraturbanken} is an initiative that is aimed at preserving Swedish literary history and culture and provides it for free to the public to use in various formats. 

\textbf{Collect and Curate Data}

Then raw data can be collected from these sources that could for example be in the form of epub files, scanned in images or readable pdfs. It is important to be careful when curating the dataset to make sure it is both reliable and comprehensive enough. In this project, the data was sourced in all of these formats and the dataset was curated to capture various aspects of swedish culture. From more factual content from Wikipedia\cite{SwedishWikipediaHuvudsida} and Museums to less factual from classic swedish literature sourced from LitteraturBanken\cite{litteraturbanken}.

\textbf{Clean and Preprocess Data}

Once raw data has been collected, it has to be cleaned and preprocessed. This includes:
\begin{itemize}
    \item Removing duplicate information.
    \item Correcting spelling mistakes or OCR (Optical Character Recognition) errors if the data format is scanned in images.
    \item Taking care of newlines, punctuation and special characters.
\end{itemize}

Well cleaned data helps with the accuracy of the RAG. For this project, all of the mentioned data formats was experimented with but in the end scanned in images was not put in the final dataset due to bad quality of the processed data. It turned out to be much simpler to process the data when OCR errors did not have to be handled. The notebooks associated with this project still contained the code that was used to experiment with these kinds of data sources.

\textbf{Chunk and Index the Data}

To make retrieval efficient and relevant, it is common to break down the text into chunks. This can be done for different chunk sizes, in the case of this project the chunk size was 200 words. These chunks can then be indexed using a vector representation and searched for using similarity search with the prompt sent by the client.

\textbf{Generating Embeddings}

In order to do RAG, the goal is to generate vector embeddings for each chunk using an embeddings model. These embeddings are supposed to capture semantic relationships between chunks of text that makes it possible for a program to retrieve relevant context from the prompt sent by the client. There are a couple of ways to do this. There are embeddings model available such as Sentence-BERT or you could use a framework to train your own model based on the data. Generally, it requires a lot of data to train a good model for generating embeddings. But is was still attempted in this project and gave some promising results in some cases. However, it was not always reliable which will be shown more in section \textit{4.4}. Therefore, a pre-trained embeddings model was also tested, \texttt{KBLab/sentence-bert-swedish-cased}. Since this project is about underrepresented languages and there might not be a model like this for all languages the goal of training one from the data was to demonstrate feasability even for other languages.

\textbf{Deploy and Maintain}

Finally, the indexed dataset could be hosted in for example a vector database so that it can be queried by the RAG system. In order to maintain it over time the dataset will require periodic updates and maintenance as the domain evolves and the capacity of the model is developed. Due to the financial limitations of this project the vector database was simulated by keeping a JSON-file of the embeddings.

\subsection{Model Selection}

The Kaggle competition does not specify a particular version of the Gemma 2 model to be used, so it was up to us to choose which one to use. First of all, the Gemma 2 model exists in different parameter sizes: 2B, 9B, and 27B. We opted for the 2B size because it balances performance and computational requirements, making it manageable to fine-tune on a single GPU while still being powerful. Furthermore, the specific model we chose was \texttt{gemma2-keras-gemma2\_instruct\_2b\_en-v1}, which is already optimized for following prompts and producing coherent outputs, as it starts from an instruction-aware checkpoint. Additionally, many other participants in the competition used this model, indicating its fit for this task. We also experimented with a Gemma 2 base model that was not fine-tuned, but it proved to be much more difficult to work with and got worse results.\\

Moving on, we chose to work in TensorFlow for several reasons. First, it’s a widely used industry standard, so learning it gives us valuable, practical experience. Second, the abundance of existing code and tutorials for fine-tuning in TensorFlow makes problem-solving easier. Finally, TensorFlow’s strong integration with TPU/JAX and other Google infrastructure offers better performance and streamlined deployment options. Also, we chose to write our code in Google Colab and run it on an A100 GPU for convenience and efficiency. Colab provides a simple environment and collaborative features enabling all of us 
to code in the same notebook. Also, the A100 GPU with 90 GB of VRAM significantly accelerates large-scale deep learning tasks compared to local hardware, enabling faster model training and experimentation.

\subsection{Fine-Tuning}
Fine-tuning refers to the process of adapting a pretrained language model to a specific task by modifying a subset of its parameters. In this project, we utilized the GemmaCausalLM model with Low-Rank Adaptation (LoRA) to efficiently fine-tune the model for our downstream task.

\subsubsection{Model Architecture and Adaptation}
We employed LoRA for parameter-efficient fine-tuning. This technique modifies a small subset of the pretrained weights using low-rank matrices as described in the formulas below:
\[
W + \Delta W = W + BA, \quad \text{where } B \in \mathbb{R}^{d \times r}, A \in \mathbb{R}^{r \times d}, \, r \ll d
\]

The advantage of LoRA is that it can efficiently adapt a pretrained model to a specific task while significantly reducing computational requirements. Unlike full fine-tuning, LoRA freezes most of our model parameters to efficiently fine tune Gemma 2 to Swedish. This can be seen in figure \ref{fig:LoRa_side_by_side} below.
\begin{figure}[H]
    \centering
    \begin{subfigure}[t]{0.5\linewidth}
        \centering
        \includegraphics[width=\linewidth]{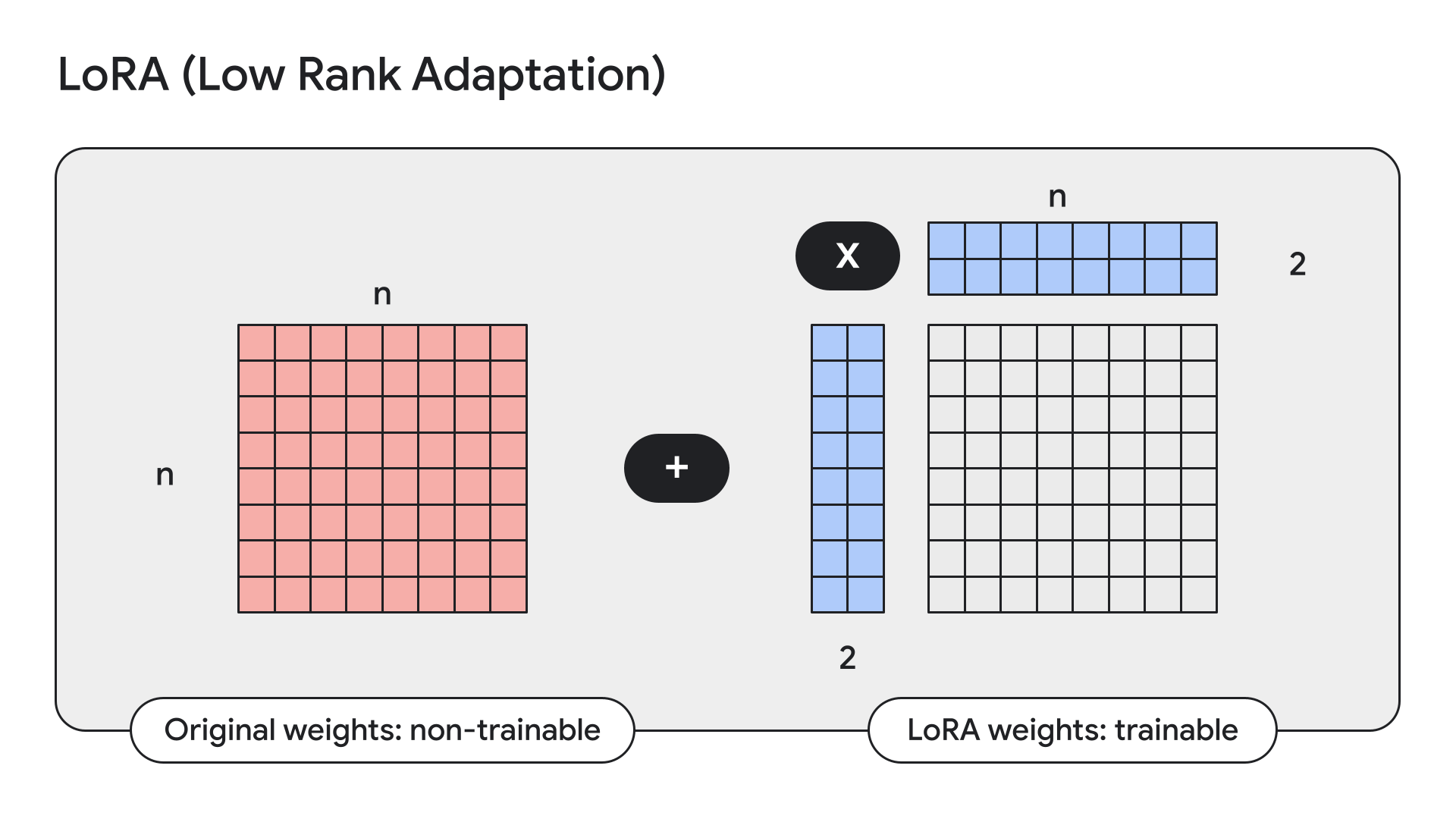}
        \caption{LoRa \cite{GoogleDevelopersBlog2024}}
        \label{fig:LoRa}
    \end{subfigure}
    \hfill
    \begin{subfigure}[t]{0.4\linewidth}
        \centering
        \includegraphics[width=\linewidth]{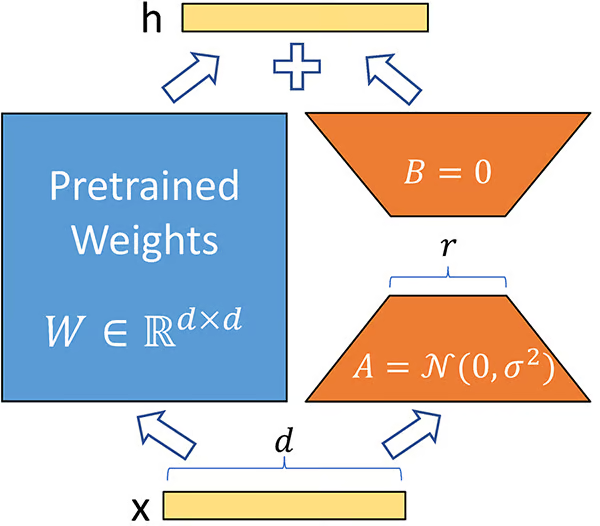}
        \caption{LoRa \cite{Anyscale2024}}
        \label{fig:LoRa_Schematics}
    \end{subfigure}
    \caption{LoRa configuration.}
    \label{fig:LoRa_side_by_side}
\end{figure}

This approach minimizes the number of trainable parameters, enabling efficient training on hardware with limited resources. As our group had access to limited resources, it was a natural choice to use this method. Additionally, LoRA facilitates faster gradient computations, which contributes to shorter training times and smoother convergence. 

\textbf{Rank Selection and Considerations}

The choice of the rank parameter \( r = 8 \) was chosen based on empirical evidence and our specific requirements. Initial experiments demonstrated that this rank provided sufficient expressiveness for the updates while maintaining the desired accuracy. Higher ranks, while potentially improving the model’s performance further, were avoided due to memory constraints. The hardware used in this project imposed strict limitations on memory utilization, and a rank of 8 balanced the trade-off between task complexity and computational efficiency. 

\begin{figure}[H]
    \centering
    \begin{subfigure}[t]{1\linewidth}
        \centering
        \includegraphics[width=\linewidth]{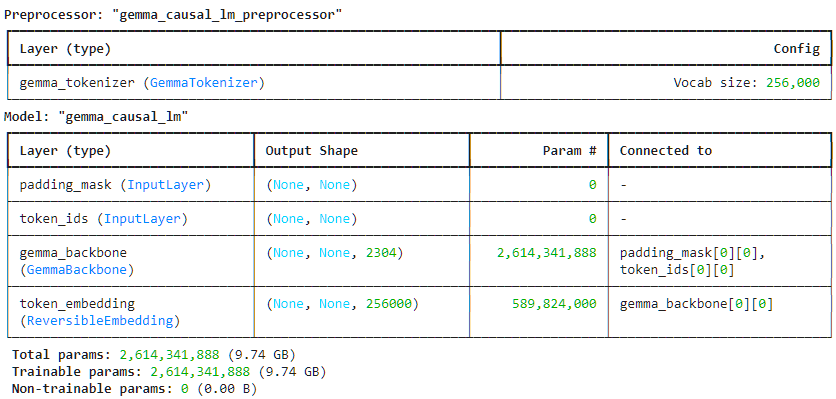}
    \end{subfigure}
    \hfill
    \begin{subfigure}[t]{1\linewidth}
        \centering
        \includegraphics[width=\linewidth]{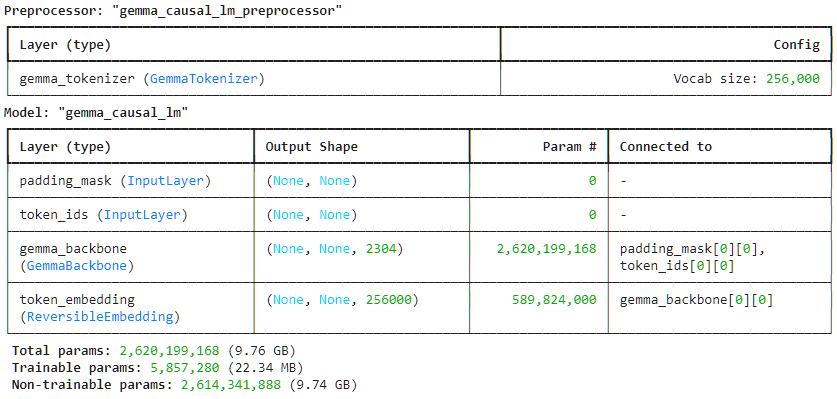}
    \end{subfigure}
    \caption{Comparison of model summaries before and after fine-tuning.}
\end{figure}
The fine-tuning process with LoRA significantly reduces the number of trainable parameters from 2.61 billion (9.74 GB) to 5.86 million (22.34 MB), highlighting its efficiency in adapting large models with minimal computational overhead.

\subsubsection{Training Setup}

The training setup for fine-tuning the GemmaCausalLM model was carefully designed to optimize the model's performance while ensuring efficient utilization of computational resources. In this subsection we will explain the data preprocessing and model configuration.

\textbf{Data Preprocessing}

To ensure robust training, the dataset was prepared by converting the formatted data into a TensorFlow Dataset object, enabling efficient handling and batching of training and validation examples. The dataset was shuffled with a buffer size equal to the total size, ensuring unbiased training by randomizing the order of examples. Prefetching, implemented using \texttt{tf.data.AUTOTUNE}, improved input pipeline efficiency by preparing data while the model processed the previous batch. Also, a train-validation split of 90\%-10\% was applied to maintain a consistent evaluation process. Both subsets were batched with a size of 16 for efficient gradient computation.

\textbf{Model Configuration}

The GemmaCausalLM model was initialized with LoRA enabled at a rank of 8. Mixed precision training was employed to leverage both float16 and float32 data types, accelerating training and reducing memory usage. Additionally, the sequence length was adjusted to 128 tokens, balancing computational efficiency and contextual understanding.

\begin{figure}[H]
    \centering
    \begin{minipage}[t]{0.4\linewidth}
        \centering
        \includegraphics[width=\linewidth]{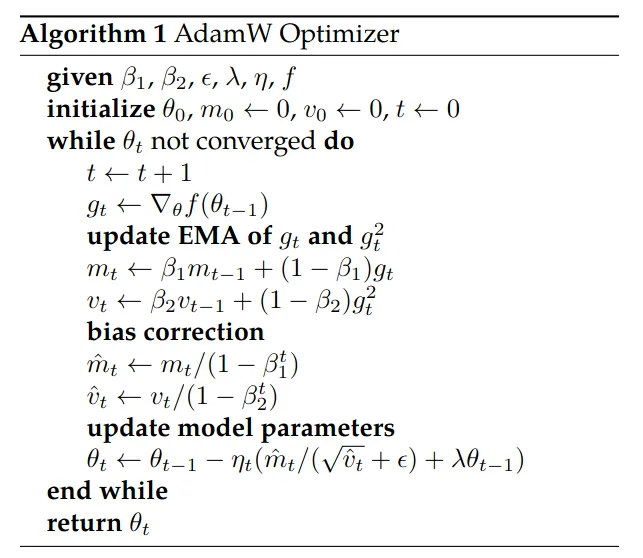}
        \caption{AdamW Algorithm \cite{LoRA2302.06675}}
        \label{fig:AdamW}
    \end{minipage}
    \hfill
    \begin{minipage}[t]{0.5\linewidth}
        \centering
        \includegraphics[width=\linewidth]{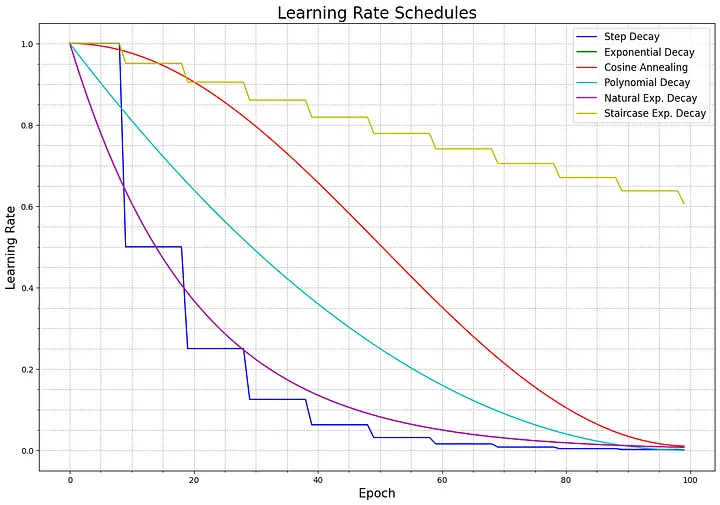}
        \caption{Learning Rate Schedulers Comparison \cite{LearningRateSchedulers2024}}
        \label{fig:CosineDecayComparison}
    \end{minipage}
\end{figure}

The optimizer used was AdamW after experimenting with Adam and Stochastic Gradient Descent as well. Furthermore, gradient clipping is also utilized to make sure the gradients do not grow too large in size. As for the learning rate, a scheduler was used. There were a couple to chose from as can be seen in figure \ref{fig:CosineDecayComparison}. In the end, the one that was chosen was to use Cosine Decay. To further optimize the training and fine-tuning we also added Warm-Up to the scheduler which makes the learning rate first gradually increase at the start of the training

To achieve efficient training and fine-tuning of the model, we employed a Cosine Decay with Warm-Up learning rate schedule. This approach ensures a gradual increase in the learning rate during the initial phase of training, followed by a smooth decay over the remaining training steps. This strategy helps stabilize the training process early on and facilitates convergence to an optimal solution.

\subsubsection{Training Workflow}

The training process was structured into a systematic workflow to ensure stability and performance improvements over epochs. Key components of the workflow included training loops, callbacks and monitoring, as seen below:

\textbf{Training Loop}

The training loop iteratively updated model parameters. The model was trained for 10 epochs with a batch size of 16, balancing computational requirements and model convergence. Sparse Categorical Accuracy was monitored on the validation set after each epoch to evaluate generalization. Mixed precision training was employed, as said before, enabling efficient scaling.

\textbf{Callbacks and Monitoring}

To ensure an adaptive and efficient training process, several callbacks were used. Early stopping was employed to halt training if the validation loss did not improve for 3 consecutive epochs, restoring the best model weights. Model checkpoints were saved after each epoch based on validation loss, ensuring that the best-performing model could be restored easily. Additionally, a \texttt{ReduceLROnPlateau} callback dynamically halved the learning rate if validation accuracy plateaued for 2 consecutive epochs. TensorBoard was integrated to log training and validation metrics, providing visual insights into model performance over epochs.

Overall, the final training process demonstrated consistent improvements over epochs, monitored through TensorBoard visualizations. Early stopping ensured optimal generalization without overfitting, while the saved model checkpoints allowed easy restoration and evaluation of the best-performing model.

\subsubsection{Training Results}
There was a lot of issues when training this model, the biggest one being overfitting. While the training loss and training accuracy steadily improves this is not the case for the validation metrics. Below in figure \ref{fig:train-metrics} it can be seen how the loss and accuracy quickly increases.
\begin{figure} [H]
    \centering
    \includegraphics[width=0.6\linewidth]{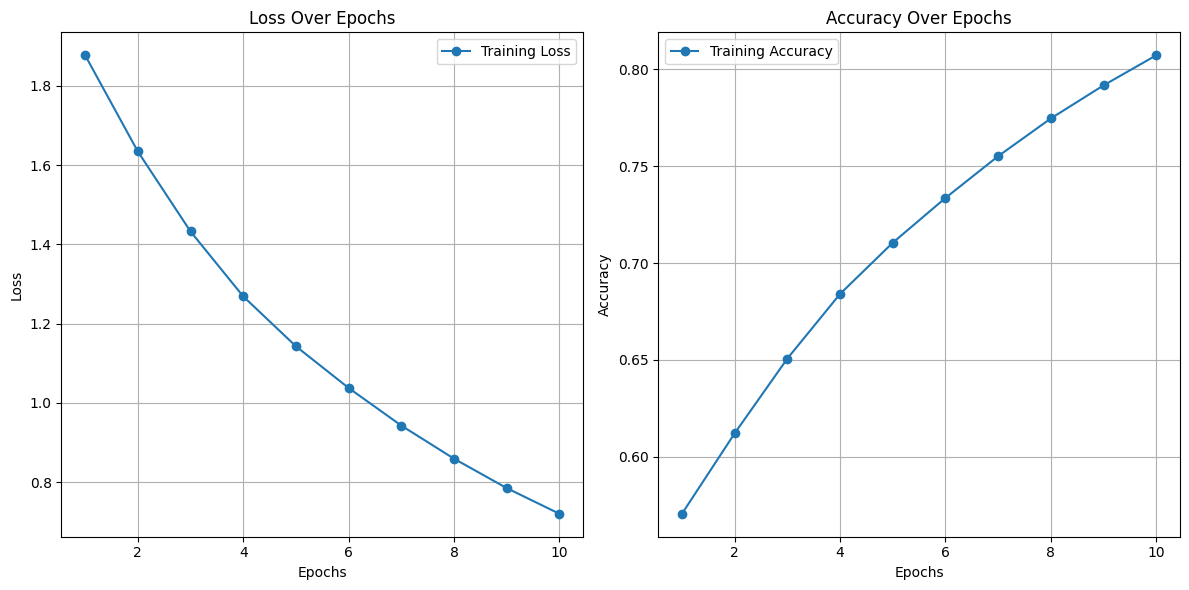}
    \caption{Training loss and accuracy}
    \label{fig:train-metrics}
\end{figure}

However, at the same time as the training metrics looks good the validation accuracy steadily drops and the validation loss keeps increasing which leads to worse results. However, after tweaking the model and implementing early stopping the following training trajectory was achieved as shown in figure \ref{fig:LossandAccuracy} below.

\begin{figure}[H]
    \centering
    \includegraphics[width=0.6\linewidth]{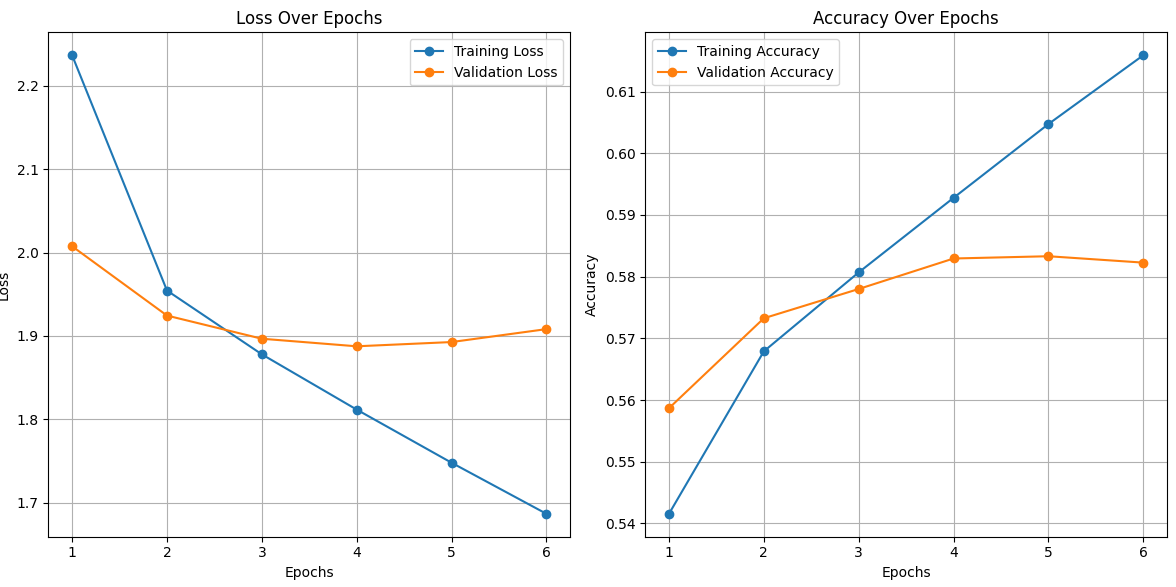}
    \caption{Training and Validation Loss and Accuracy}
    \label{fig:LossandAccuracy}
\end{figure}

Below in figure \ref{fig:la} is another example of another configuration with a lower learning rate that also ended up overfitting. We address this result in details in \hyperref[sec:interpretation]{here}
\begin{figure}[H]
    \centering
    \includegraphics[width=0.6\linewidth]{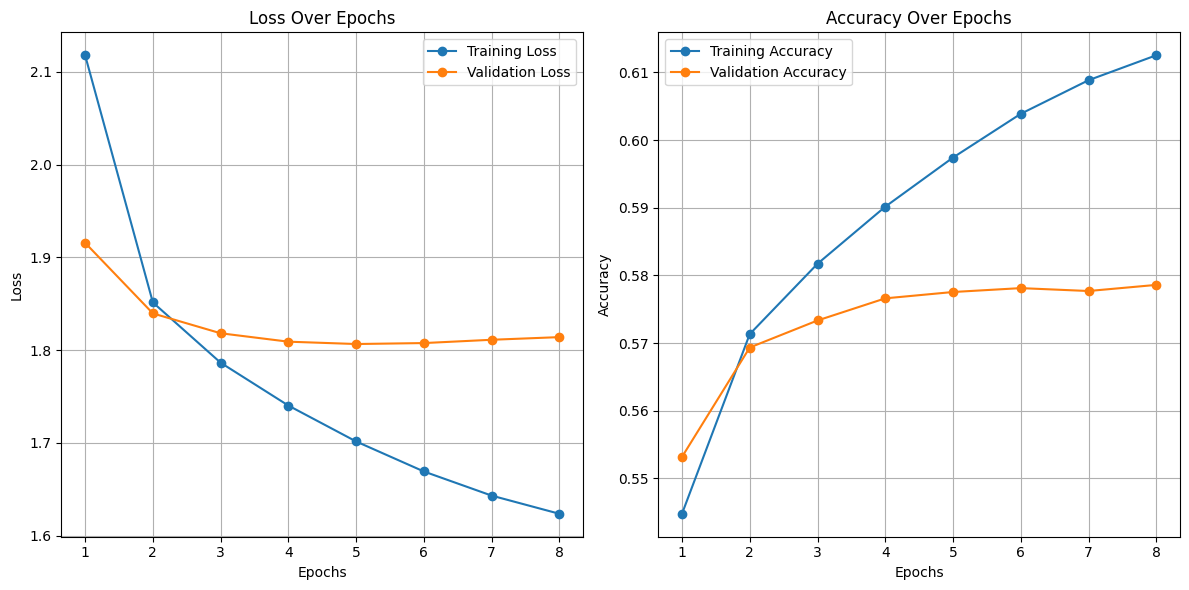}
    \caption{Training and Validation Loss and Accuracy}
    \label{fig:la}
\end{figure}

\subsection{Retrieval Augmented Generation}

Retrieval Augmented Generation (RAG) is the process of optimizing the outputs of a LLM by utilizing a knowledge base outside of the models training data before generating the response to a prompt according to \cite{AWSRetrievalAugmentedGeneration}. LLMs are already trained on large volumes of data and the goal of using RAG is to extend the capabilities of the model to specific domains without the need to retrain the model\cite{AWSRetrievalAugmentedGeneration}. It can be expensive to retrain these large models so RAG offers an alternative. According to \cite{AWSRetrievalAugmentedGeneration}, it is a cost-effective approach to improve the responses of a LLM so that is is relevant and accurate in various contexts.

Because of this it was decided that it could potentially be a good fit for this project where Gemma 2 was to be optimized for Swedish language and culture. According to \cite{AWSRetrievalAugmentedGeneration}, RAG gives greater control over the generated text output and it becomes easier to understand why the model gave the response that it did since it is possible to look at the context retrieved. 

One example from using the Gemma 2 model before fine-tuning and RAG shows how it hallucinates and generates false outputs which further motivates the use of RAG. It is shown below
\begin{verbatim}
    Prompt: Vem är Sveriges statsminister?
    Prediction: 
    Vem är Sveriges statsminister?
    
    Sverige har en parlamentarisk demokrati med en statliga president som är 
    den högsta ledaren i landet. Presidenten utser också statsminister.
    
    Presidenten utser statsminister från den största partiet i riksdagen. 
    
    Statsministerns roll är att leda regeringen och att utföra statliga 
    uppdrag.
    
    
    Expected Response: 
    Vem är Sveriges statsminister? 
    
    Sveriges nuvarande statsminister är Ulf Kristersson, som tillträdde 2022.
\end{verbatim}

In the example above the Gemma 2 model generates a response that says Sweden is a Republic, which is not the case since it is a Monarchy without a president. It clearly needs more information about Sweden. It also does not answer the question about who is the prime minister and is quite bad at Swedish grammar, which will be tackled with fine tuning.

The RAG process involves three primary steps that are explained in \cite{Lewis2020RAG}.

\textbf{Context Retrieval} 

Given an input prompt \( q \), the program should retrieve \( k \) relevant documents or chunks or sentences \( \{d_1, d_2, \ldots, d_k\} \) from a base of knowledge \( D \). This process can be expressed as:
    \[
    \{d_1, d_2, \ldots, d_k\} = \arg \max_{d \in D} \text{Score}(q, d),
    \]
    where \( \text{Score}(q, d) \) is the measure of similarity used in the project. The similarity in this case is between the embeddings of the query \( q \) and a document/chunk/sentence \( d \), depending on what type of data is used.

\textbf{Combination with original prompt}

The retrieved context is concatenated with the original input query or prompt \( q \) to form a new query with the context and the original prompt \( c \). The combined context and original prompt is then sent to the LLM:
    \[
    c = \text{concat}(q, d_1, d_2, \ldots, d_k).
    \]

\textbf{Response Generation}

The LLM then generates a response \( r \) that utilizes the combined context \( c \):
    \[
    r = \text{LLM}(c).
    \]

In this project, the similarity score \( \text{Score}(q, d) \) that was used was cosine similarity between the query and the embeddings. 

The final response \( r \) is in other words influenced both by the quality of the context that was retrieved and by the LLM's ability to give good responses and interpret the combined query or prompt \( c \).

In this project, RAG was implemented using the dataset that was sourced as described in section \textit{4.1.2}. Then the method of generating the embeddings was not as straightforward. Two approaches were tested, using a pretrained model to generated the embeddings and training a new model using a framework like FastText. 

\subsubsection{FastText Model}
According to \cite{FastText2024}, FastText is an open-source, free and lightweight library that can be used to learn text representations. It is included in gensim\cite{rehurek_lrec} and can easily be used to train a model that can generate embeddings as shown below. More information about the FastText implementation in Gensim can be found at \cite{GensimFastText}.

\begin{verbatim}
    model = FastText(flattened_sentences,
        vector_size=200, 
        window=7, 
        min_count=1, 
        workers=4, 
        sg=1, 
        epochs=10)
        
    model.save("fasttext_model.model")
\end{verbatim}

Below is an example of this model after it had been trained on data from the RAG dataset. It is important to note that in order for it to work well. The data has to be properly processed an tokenized beforehand as explained in section \textit{4.1.2}. 

\textbf{Prompt:}\\
Hur ser Sveriges kultur ut? Svara med max 10 meningar.

\textbf{Context as a small list of sentences:}

['Sveriges rikes lag.', 'En studie av materiell kultur och ekonomi hos Sveriges första fångstfolk.', 'Ordet neutralitet försvann från Sveriges säkerhetspolitiska linje 2007.', 'Amerikansk kultur har sedan länge ett stort inflytande.', 'Svensk kultur är vidare starkt individualistisk och antinationalistisk, även självk ritisk.', 'Sveriges statsminister Ulf Kristersson.', 'Leve Sverige!', 'Till Sverige kom relikvariet 1632 som krigsbyte under trettioåriga kriget.', '[138] Svensk kultur är starkt egalitär och mycket öppen mot omvärlden.', 'Istället handlade den om Sverige idag: Sverige är ett demokratiskt land med fred, frihet och respekt för mänskliga rättigheter, oavsett religion eller ursprung.']

The prompt translates to "What does the culture of Sweden look like? Answer with at most 10 sentences." and it can be seen that many relevant sentences was retrieved using this FastText model. For example:

\texttt{'Amerikansk kultur har sedan länge ett stort inflytande.'} roughly translates to \texttt{'American culture has since long ago had a large influence.'}.

\texttt{'Svensk kultur är starkt egalitär och mycket öppen mot omvärlden.'} roughly translates to \texttt{'Swedish culture is strongly egalitarian and very open to the outside world.'}.

\texttt{'Svensk kultur är vidare starkt individualistisk och antinationalistisk, även självk ritisk.'} roughly translates to \texttt{'Swedish culture is furthermore strongly individualistic and anti-nationalistic, as well as self-critical.'}.

These are all good examples but there are also situations where the model does not generate good outputs at all. Therefore a pretrained model was also experimented with. Due to the nature of this project where the goal is to focus on underrepresented languages this could be considered a bad idea since not all languages have access to pretrained models. However, since it has been shown that FastText or similar models can be trained, it was decided that exploring the use of pretrained models was a good idea to see what performance the RAG could generate with a model trained on more data. In order to make the FastText model more feasible more data would have to be used to train it.

\subsubsection{Pretrained Sentence-BERT}
The pretrained model used for this can be initialized in python using the following code:
\begin{verbatim}
    model = SentenceTransformer("KBLab/sentence-bert-swedish-cased")
\end{verbatim}

The pretrained model did, as expected, give more interesting results especially since it utilized chunking to not retrieve individual sentences but entire chunks. Below is an example when giving the following sentence to the embeddings model: "Hur ser Sveriges kultur ut?".

\textbf{Chunk:}
\begin{quote}
Tätorter från 31 december 2020, kommuner från 30 september 2024) Alfred Nobel, instiftare av Nobelpriset. Muslimerna uppskattas 5 procent, eller 25 000, vara praktiserande (i betydelsen att de deltar i fredagsbönen och ber fem gånger om dagen). Det finns även buddhister, judar, hinduer och bahá'íer i Sverige. Bland de övriga finns enstaka som praktiserar modern asatro och traditionell samisk religion. Svensk kultur är en del av de nordiska, germanska och västerländska kulturområdena. Svenska kulturuttryck inom konst, musik och litteratur ansluter sig främst till dessa traditioner. Det offentliga stödet till kulturen är omfattande i Sverige. Folkligt deltagande är stort inom många kulturverksamheter, till exempel körsång, som engagerar tiotusentals svenskar. Värderingsmässigt skiljer sig svensk kultur starkt från världens genomsnittsvärderingar genom att vara mycket mer universalistisk, sekulär och inriktad mot postmaterialistiskt självförverkligande. Svensk kultur är starkt egalitär och mycket öppen mot omvärlden. Amerikansk kultur har sedan länge ett stort inflytande. Svensk kultur är vidare starkt individualistisk och antinationalistisk, även självkritisk. Maximal jämställdhet mellan kvinnor och män har ett centralt värde.
\end{quote}

\textbf{Similarity Score:} 0.7748

\textbf{Metadata:}
\begin{itemize}
    \item \textbf{Title:} Sverige – Wikipedia
    \item \textbf{Document ID:} 1\_chunk54
    \item \textbf{File Path:} \texttt{data\_preprocessing/data/wikipedia/Sverige.pdf}
    \item \textbf{Subdirectory:} \texttt{data\_preprocessing/data/wikipedia}
\end{itemize}

It can clearly be seen that this response is more interesting. By using chunks of data more information will also be included and with this new approach such as which document the information came from.

\subsubsection{RAG Architecture}
This section will cover how the RAG mechanic was integrated with the Gemma 2 model in the notebooks used for this project. First of all when using the FastText model the problems with the model being trained on not enough data became painfully apparent. When giving the embeddings model a random question the embeddings model will not give back a good context, instead it will just give back other questions which is not helpful. Therefore the following method for utilizing it was implemented as shown below in figure \ref{fig:sentence-rag}.

\begin{figure} [H]
    \centering
    \includegraphics[width=0.75\linewidth]{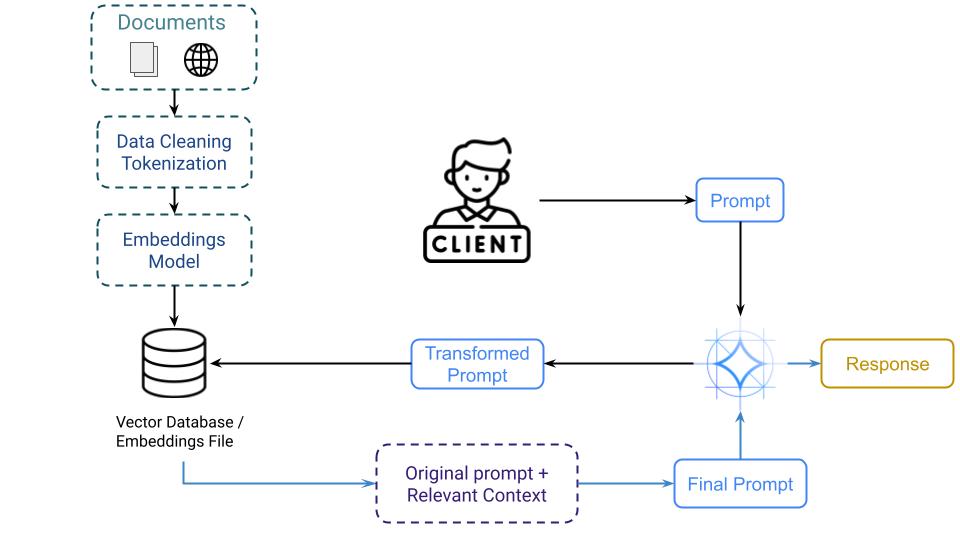}
    \caption{Rag Architecture using FastText}
    \label{fig:sentence-rag}
\end{figure}

First of all, as can be seen in figure \ref{fig:sentence-rag}, the dataset is used to train an embeddings model and to generate embeddings stored in a Vector Database or an Embeddings File. In order to work around the problem with the FastText model trained on limited data, an alternative pipeline to the client sending the prompt directly to the model was implemented. How it works is that when the client sends a prompt the Gemma 2 model is first asked to transform the prompt into a simple statement that fits well with the embeddings and embeddings model. It is done with the following instructions: 
\begin{verbatim}
    context_prompt = f"Översätt inte detta. 
                       Gör bara om det från en fråga till ett statement:
                       '{og_prompt}'. Answer with at most 4 words."
\end{verbatim}  

This basically tells the model to not translate the prompt from Swedish to english which was a big problem that was encountered. It also tells the model to rewrite the prompt from a question into a statement and to answer with at most 4 words. That transformed prompt, that is not a question, is then used to retrieve context from the embeddings file or vector database and the context is combined with the original prompt to get the final prompt. This final prompt is then given to the Gemma 2 model that generates the response. This method proved to be quite successful. One example is described below, without using this approach and passing the prompt directly to the embeddings and using the prompt: "Hur ser Sveriges kultur ut? Svara med max 10 meningar." ("What does the culture of Sweden look like? Answer with at most 10 sentences."). While some of the context is alright for this specific prompt, there are also questions in the retrieved context that have nothing to do with the desired answer such as "Hur ﬁras dagen?" ("How is the day celebrated?"). By first passing it to the Gemma model to string sent to the embeddings for retrieval is "Sveriges kultur" which improves the context retrieved significantly.

In the case where a pretrained model is used for generating embeddings the process is simplified since there is no need to pass the prompt to the model twice and transform it to get good responses from the model. The method of using it in that case is shown below in figure \ref{fig:simplified-rag}.

\begin{figure} [H]
    \centering
    \includegraphics[width=0.75\linewidth]{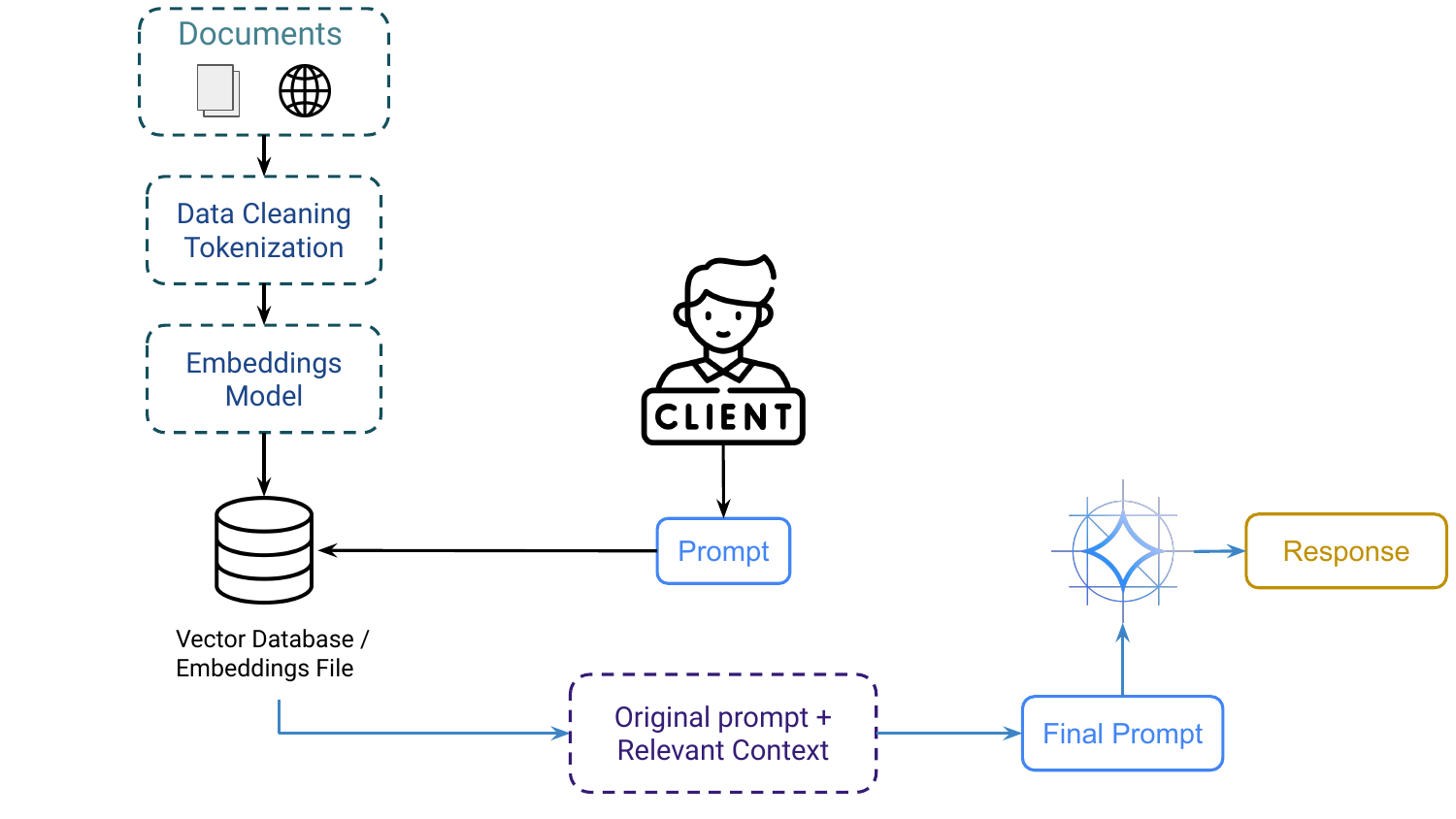}
    \caption{Rag Architecture using Sentence-BERT}
    \label{fig:simplified-rag}
\end{figure}

The pretrained model is trained on more data and better att identifying similarity between text which is why the results are much better. It is therefore not necessary to send the prompt and have Gemma 2 transform it, instead it can be sent directly to retrieve context.

\section{Results and Evaluation}
\subsection{Evaluation Metrics}

The fine-tuned GemmaCausalLM model was evaluated using multiple metrics to assess key performance aspects: accuracy, linguistic quality, and semantic fidelity. These metrics were selected to align with common LLM tasks, including question answering, summarization, and machine translation.

\subsubsection{Quality of Question Answering and Span-Based Tasks}

For question answering and span-based tasks, we used \textbf{Exact Match (EM)} and \textbf{F1 score}. EM measured the percentage of predictions that perfectly matched the ground truth, ensuring strict correctness. F1 provided a balanced measure of precision and recall, allowing partial matches to be evaluated comprehensively. Together, these metrics assessed both the precision and completeness of the model's outputs.

\subsubsection{Quality of Summarization}

To evaluate summarization tasks, we employed \textbf{ROUGE} metrics, which measured how well the generated summaries preserved the content and structure of the reference texts. By comparing sequences of words and phrases, ROUGE provided insights into the model’s ability to retain key information while maintaining coherence.

\subsubsection{Quality of Translation and Text Generation}

For translation and text generation, we used BLEU, METEOR, BERTScore, and COMET:

\begin{itemize}
    \item \textbf{BLEU} measured the overlap of n-grams between the generated text and reference, emphasizing fluency and adequacy.
    
    \item \textbf{METEOR} extended BLEU by incorporating synonyms, stemming, and paraphrasing into its evaluation, providing a more nuanced measure of linguistic quality and alignment with reference texts.
    
    \item \textbf{BERTScore} used contextual embeddings to evaluate semantic similarity, ensuring the generated text preserved meaning, even when rephrased.
    
    \item \textbf{COMET} focused on cross-lingual translation, assessing how well the model maintained adequacy and fluency in multilingual contexts.
\end{itemize}

\subsubsection{Comprehensive Evaluation Across Tasks}

For question answering, EM and F1 scores captured accuracy and completeness. In summarization tasks, ROUGE measured content preservation and coherence. For translation and text generation, BLEU and METEOR assessed fluency and adequacy, while BERTScore and COMET ensured semantic and cross-lingual fidelity. This comprehensive evaluation framework ensured that the fine-tuned model was rigorously assessed across a diverse range of tasks, balancing strict accuracy with semantic and linguistic quality.

\subsection{Results}
\subsubsection{Results for Question Answering and Span-Based Tasks}
The model's performance on the question answering and span-based tasks is summarized in Table~\ref{tab:qa-results}. The EM and F1 Score demonstrate the model's ability to generate accurate and complete answers.

\begin{table}[h!]
    \centering
    \begin{tabular}{|c|c|c|c|}
        \hline
        Metric & Pretrained Model & Fine-Tuned Model & Fine-Tuned Model with RAG \\
        \hline
        EM     & 0\%           & 0\%           & 0\% \\
        F1     & 64.98\%           & 47.72\%           & 77.63\% \\
        \hline
    \end{tabular}
    \caption{Question Answering Results}
    \label{tab:qa-results}
\end{table}

It is worth mentioning that the EM and F1 metrics, while widely used for evaluating extractive QA models, may not be the most suitable when prioritizing creativity and diversity. These metrics mainly focus on how closely the answers match the reference text and how accurate they are at a surface level, potentially undervaluing responses that are semantically correct but phrased differently or reflect nuanced reasoning. Additionally, achieving high scores with these metrics often depends significantly on the specificity of the prompt, as well-crafted, precise prompts guide the model toward extracting the exact spans required for evaluation. Alternative metrics that better capture semantic similarity and contextual relevance may be more appropriate for other QA tasks where flexibility is valued.

\subsubsection{Results for Summarization}
For summarization, ROUGE metrics were used to evaluate the quality of the generated summaries. Table~\ref{tab:summarization-results} shows the scores for ROUGE metrics, indicating the model's effectiveness in retaining key information and producing coherent summaries.

\begin{table}[h!]
    \centering
    \begin{tabular}{|c|c|c|c|}
        \hline
        Metric     & Pretrained Model & Fine-Tuned Model  \\
        \hline
        ROUGE-1   & 0.69            & 0.76      \\
        ROUGE-2   & 0.64           & 0.71   \\
        ROUGE-L   & 0.68            & 0.74   \\
        \hline
    \end{tabular}
    \caption{Summarization Results}
    \label{tab:summarization-results}
\end{table}

\subsubsection{Results for Translation and Text Generation}
The model's performance in translation and text generation tasks was evaluated using BLEU, METEOR, BERTScore, and COMET. The results in Table~\ref{tab:translation-results} highlight its fluency, semantic fidelity, and multilingual adequacy.

\begin{table}[h!]
    \centering
    \begin{tabular}{|c|c|c|c|}
        \hline
        Metric     & Pretrained Model & Fine-Tuned Model  \\
        \hline
        BLEU       & 0.29            & 0.46    \\
        METEOR     & 0.82            & 0.51   \\
        BERTScore  & 0.77            & 0.76   \\
        COMET      & 0.60            & 0.72   \\
        \hline
    \end{tabular}
    \caption{Translation Results}
    \label{tab:translation-results}
\end{table}

\subsubsection{Interpretation and conclusion}
\label{sec:interpretation}
The results demonstrate strong performance across a lot of task, particularly in translation, where the model achieved high BLEU and BERTScore values, indicating fluency and semantic fidelity. 
The lower METEOR score observed for the fine-tuned model may suggest that the model has overfitted to the fine-tuning dataset. This overfitting implies that while the model performs well on the specific data it was trained on, it struggles to generalize effectively to unseen examples.

The model also demonstrated good performance in question answering, as highlighted by a higher F1 score. However, the EM score is always 0, indicating no perfect alignment between predicted and ground-truth answers. While this result means that there is no precise textual alignment, it is less critical in certain applications compared to semantic accuracy. While summarization scores were satisfactory, there is potential for improvement in capturing nuanced information.

\paragraph{Quality of datasets}
The effectiveness of fine-tuning a model is directly influenced by the quality of the dataset used. In this project, the dataset of crafted prompt-response pairs played a crucial role in determining the performance of the fine-tuned model. However, the results, including lower METEOR and F1 scores, suggest that the dataset may lack the diversity and comprehensiveness needed for broader generalization.

The dataset appears to be too narrow, limiting the model's ability to handle prompts and responses that deviate from its training data. This lack of variability impacts the model's flexibility and hinders its potential to generate contextually relevant and semantically rich responses.

We address these limitations more in \hyperref[sec:future-work]{Section 6} and \hyperref[sec:future-work]{Section 7}.

\subsubsection{Subjective evaluation of the result}
The quality of a response generated from a prompt can be a very subjective thing. Given two responses and two observers, the observers might prefer different responses. From this idea it could be argued that using subjective evaluation methods are relevant. In this project, the subjective evaluation was carried out by the group members with a Swedish background. From the observations made and with the knowledge the evaluators posses as native Swedish speakers, the methods proposed in this project do improve the responses.

As an example using the FastText model that was trained specifically for this project. To the question: "Hur ser Sveriges kultur ut? Svara med max 10 meningar.", the model gave a pretty uninteresting and not representative answer according to the group. After fine tuning and using RAG the answer was subjectively much more interesting and accurate. The before and after response is shown below:

\textbf{Before:}
\begin{quote}
Sverige har en rik kulturhistoria med många olika traditioner och influenser. 

**Folkmusik:** Sverige har en stark tradition av folkmusik, med instrument som violin, accordion och fiol. 

**Konst:** Sverige har en lång tradition av konst, med många framstående konstnärer som Carl Larsson, Edvard Munch och Anders Zorn. 

**Litteratur:** Sverige har en rik litteraturhistoria, med många framstående författare som Astrid Lindgren, Selma Lagerlöf och Stieg Larsson. 

**Arkitektur:** Sverige har en unik arkitektur, med många vackra byggnader från olika epoker. 

**Mat:** Sverige har en rik matkultur, med många traditionella rätter och delikatesser. 

**Festivaler:** Sverige har många populära festivaler, som Melodifestivalen och Stockholm Pride. 

**Natur:** Sverige har en fantastisk natur, med många vackra skogar, sjöar och öar. 

**Idrott:** Sverige har en stark idrottskultur, med många framstående idrottsmän. 
\end{quote}

\textbf{After:}
\begin{quote}
Sveriges kultur är en komplex och mångfacetterad konstruktion som präglas av en rad olika faktorer. Den är starkt influerad av svensk historia, med en lång tradition av folkmusik, konst och litteratur. Sverige har en stark tradition av egalitär och öppenhet, vilket syns i dess starka socialtjänst och välfärdssystem.  Sveriges kultur är också starkt individualistisk och antinationalistisk, vilket gör det till en unik och dynamisk kultur.  Sveriges kultur är också starkt präglad av den svenska naturen, med många vackra parker och skogar.  Sveriges kultur är också starkt influerad av den svenska traditionen av folkmusik, konst och litteratur.  
\end{quote}

Not only is the format more what would be expected. The model is able to incorporate information from external sources using RAG which can be seen in how the new response captures that Sweden is a individualistic and anti-nationalistic country with a strong welfare system.

The context provided by the pretrained embeddings model was even better which shows the potential of this approach.

\section{Discussion, Challenges, and Solutions}
\label{sec:discussion}
This section will cover some discussion about the implementation of the different aspects of this project as well as how things could have been done differently.

\subsection{The impact of the constraints on this project}
An interesting point during one lecture was that the relevance of the work increases when working with a larger model. This is because even with all the optimization that is done to a small model, the much larger model is probably going to be better still. Furthermore, it is possible that the techniques used to improve a tiny model does not scale when working with a larger model. Due to the financial and computational constraints of this project, the model used was a small version of Gemma 2 which is for these reasons not ideal. If a larger model would have been used, the chances are that it would perform better and that other methods for improving performance could have been used. It is also a possibility that the same methods could still be used.

\subsection{Difficulty in comparing models}
The responses generated by a model are inherently evaluated by the individual who wrote the prompt. Since human perspectives vary, this evaluation is subjective. As previously mentioned, Reinforcement Learning from Human Feedback can be implemented to improve results by incorporating subjective measures. This allows the model to generate responses tailored to user preferences, such as text length, structure, and language style.

The evaluation of the models in this project was also carried out on a set of test prompts and answers which means that it is possible that it would perform worse on a different set of tests. To get a fair comparison to other models, they would have to be tested on the same test cases.

\subsection{The impact of future computing costs}
One of the reasons for working with RAG, as pointed out in \cite{AWSRetrievalAugmentedGeneration} and section \textit{4.4}, was the cost of retraining large models. As RAG can be a cost effective way of improve the responses of a LLM it was seen as a good choice since the people working to fine tune a LLM for their domain might not have access to a lot of resources.

However, as Nvidia CEO Jensen Huang says in \cite{NvidiaAI2024}, the cost of the computation for AI will go down and if that is the future, which could likely be the case, the relevance of using RAG might go down. If the cost of compute goes down faster than the amount of data available then in the future it might not be necessary to utilize RAG. But, as the cost of compute decreases other bottlenecks might appear such as the quality of the data the models are trained on and how the data is curated. To use RAG to instead utilize external data in a more dynamic way might be valuable even if the cost of compute goes down. Another strength of RAG is the explainability of the results. It becomes easier to see why the model gave a response and where the data came from, which is crucial for building trust in AI systems and ensuring their outputs are reliable and transparent. A topic highly relevant right now. 

\subsection{RAG and Fine Tuning}

As can be seen in figure \ref{fig:method-matrix}, there seems to be logical reasons as to why doing both RAG and Fine Tuning can be a good idea. As shown in section \textit{4.4}, the base pretrained model performs poorly in Swedish. The language is off and it does not know the facts about Sweden. Therefore doing both was a natural choice since there was a need for external knowledge and adaptation of the model to the language. 

More work is needed both for fine-tuning and RAG, especially when it comes to the datasets. For fine-tuning, we ultimately decided to use 500 prompts and responses. While this may be slightly fewer than what others have used online, this approach combined with RAG still produced good results. However, if better fine-tuning performance is desired, a significantly larger dataset would be required. A larger dataset would allow the model to learn a more comprehensive representation of the target language and domain-specific knowledge, ultimately improving its ability to generate accurate and contextually appropriate responses. And while a lot of text was used for RAG, to keep a model that is up to date and accurate a more comprehensive dataset would have to be crafted.

\section{Conclusion and Future Work}

\subsection{Conclusion}

In summary, this project has demonstrated that by combining LoRA and RAG, it is feasible to adapt Gemma 2 for an underrepresented language like Swedish. The results show improvements in the tasks question answering, summarization, and translation tasks. RAG proved particularly effective in dynamically supplying the model with up to date information and reducing off-topic responses.

However, there is a lot of room for improvement. With more time and testing more configurations for the training perhaps the performance could have been greater. If the struggles with overfitting, resource limitations and small size of the datasets was fixed. The result could have been much better.

\subsection{Future Work}
\label{sec:future-work}
The dataset used for fine tuning and RAG in this project are relatively small. In order to achieve better results larger datasets that are more carefully crafted could be utilized to better capture complex nuances in the language that is being worked on.

Furthermore, it was shown that a pretrained embeddings model was much better at capturing semantic relationships and it begs the question if there is room for further improvement. By using vast amounts of data in the target language to create a better embeddings model the performance could be increased.

Another interesting topic to explore would be to do reinforcement learning with human feedback. This is something that \cite{huggingface_gemma_jpn} did when fine-tuning Gemma 2 for Japanese. This would allow the model to become better at understanding what constitutes a good and relevant response which could increase the performance.

Moreover, the evaluation of the model was carried out on a relatively small number of prompt-response pairs and only two people who served as the judges for the relevance of the responses. When completing a scaled up version of this project, involving more judges and a more structured way to evaluate the responses in comparison to other models would be a good idea. 

By improving on these suggestions, Gemma 2 could increasingly serve diverse cultural contexts and preserve global language heritage. \\

On a personal note, the group is satisfied with the results of this project and have learned a lot.

\newpage
\section{References}
\bibliographystyle{plainnat}
\bibliography{main.bib}

\end{document}